\documentclass{article}

\usepackage[preprint]{neurips_2025}

\usepackage[utf8]{inputenc}
\usepackage[T1]{fontenc}
\usepackage[hidelinks]{hyperref}
\usepackage{url}
\usepackage{booktabs}
\usepackage{multirow}
\usepackage{amsfonts}
\usepackage{amsmath}
\usepackage{amssymb}
\usepackage{mathtools}
\usepackage{microtype}
\usepackage{xcolor}
\usepackage{graphicx}
\usepackage{cleveref}

\newcommand{\R}{\ensuremath{\mathbb{R}}}

\DeclareMathOperator{\topk}{TopK}

\newcommand{\URef}{\ensuremath{U_{\mathrm{ref}}}}

\newcommand{\Encoder}{\ensuremath{W_{\mathrm{enc}}}}
\newcommand{\Decoder}{\ensuremath{W_{\mathrm{dec}}}}
\newcommand{\Mdict}{\ensuremath{M}}
\newcommand{\Ksparse}{\ensuremath{k}}

\title{Post-Hoc Sparse Coding of Latent Communication\\Between Vision-Language Model Agents}

\author{
  Di Wu \\
  Xi'an Jiaotong-Liverpool University \\
  \texttt{Di.Wu23@student.xjtlu.edu.cn}
  \And
  Xiaohui Zhu \\
  Xi'an Jiaotong-Liverpool University \\
  \texttt{Xiaohui.Zhu@xjtlu.edu.cn}
}

\begin{document}

\maketitle

\begin{abstract}
Latent-space communication allows heterogeneous vision-language model agents to exchange continuous representations without serializing visual and reasoning states into text.
Vision Wormhole realizes this approach by translating visual features into a universal latent representation that can be consumed by another model, but every message is transported as a dense tensor of the same size regardless of its content.
A fixed-capacity dense tensor therefore need not have a fixed effective information density: some messages may use only a small fraction of the available representational degrees of freedom.
This observation suggests that the communication channel may be substantially compressible.
We study its redundancy by fitting a post-hoc sparse autoencoder to frozen Vision Wormhole activations and measuring reconstruction, downstream utility, feature reuse, and token-level interventions across nine reasoning benchmarks.
Relative to the original float32 transport, a uint16-index/float16-value sparse payload with $k{=}4$ active coefficients per token reduces the transmitted bytes by $128\times$.
In a single-run evaluation, the seven-task non-AIME mean accuracy changes from 49.85\% to 49.77\%.
The fitted 4096-element dictionary uses only 50 features, and task-level active sets have a mean pairwise Jaccard similarity of 0.906.
These measurements establish strong post-hoc compressibility relative to the original transport, but do not yet isolate the incremental contribution of sparse coding from position selection, reduced precision, low-rank structure, or SAE optimization effects.
The results motivate matched-payload comparisons and communication mechanisms whose payload adapts to the information used by each message.

\end{abstract}

\section{Introduction}
Cooperative vision-language agents must communicate visual observations and intermediate reasoning across models.
Natural-language messages provide a convenient interface, but they require the sender to serialize a continuous multimodal state into discrete text before the receiver can act on it.
Latent-space communication avoids this conversion by exchanging continuous representations directly, preserving information that may be difficult to express in language.

Vision Wormhole~\citep{VisionWormholeBaseline} provides a practical solution for heterogeneous agents.
It uses a visual encoder and a learned codec to translate model-specific visual features into a universal latent representation $\URef \in \R^{N \times D}$ that can be injected into a different receiver model.
This design enables direct communication across incompatible representation spaces, but its transmitted tensor has a fixed shape.
Every message therefore uses the same nominal transport capacity, even when its content and informational demands differ.

The informational demand of a message, however, can vary with its visual content, task, and reasoning stage.
A fixed-capacity dense tensor therefore need not have a fixed effective information density: some messages may use only a small fraction of the available representational degrees of freedom.
We do not claim to estimate Shannon entropy directly.
Instead, we ask an operational question: how much of the dense Vision Wormhole channel is required to reconstruct its messages and preserve their downstream utility?
Answering this question is a necessary step before designing adaptive or training-time communication mechanisms.

We investigate the channel with a post-hoc sparse autoencoder (SAE) trained on frozen Vision Wormhole activations.
Because neither the agents nor the original codec is updated, the SAE exposes structure already present in the learned channel rather than imposing sparsity during its training.
We use it as both a measurement instrument and a compressor: sparse reconstruction measures effective representational redundancy, task accuracy tests whether the retained structure remains useful to the receiver, and feature analyses examine whether that structure changes across tasks, token roles, and reasoning rounds.

The experiments show that the original dense transport admits a compact post-hoc code, achieving a $128\times$ payload reduction with little change in single-run macro-average accuracy.
Feature analyses further indicate a small, broadly reused support and token-role-dependent activation patterns.

Our contributions are:
\begin{enumerate}
    \item We formulate and empirically study the mismatch between Vision Wormhole's fixed-shape latent payload and the empirical structure used by its communication channel.
    \item We measure an end-to-end $128\times$ payload reduction relative to the original float32 transport and test the reconstructed messages on downstream tasks under explicit payload-accounting assumptions.
    \item We characterize the support learned by a post-hoc SAE across nine tasks, quantifying its size and cross-task reuse.
    \item We analyze token-group interventions to distinguish shared feature support from role-dependent activation patterns.
\end{enumerate}

These findings also motivate future work on matched-payload baselines, behavioral interventions, cross-pair validation, and communication protocols with message-dependent sparse budgets.

\section{Related Work}
\subsection{Latent Communication Between Heterogeneous Agents}

Communication is a central design choice in multi-agent systems.
Language, tool calls, and hand-designed discrete protocols are easy to route between agents, but they constrain what one model can communicate to what can be serialized explicitly.
Learned communication in cooperative multi-agent reinforcement learning instead allows agents to develop differentiable or discrete protocols, as in DIAL/RIAL and emergent grounded languages~\citep{Foerster2016DIAL,Mordatch2018Emergence}.
For heterogeneous vision-language models, the challenge is sharper because their internal visual representations are not directly compatible.
Vision Wormhole~\citep{VisionWormholeBaseline} addresses this problem with a learned codec that maps a sender's visual hidden states into a universal latent representation and injects it into a receiver.
It establishes that latent communication can bridge heterogeneous models.
Our study asks how efficiently the resulting channel uses its fixed representational capacity.

Recent work has also compared text, dense-latent, and SAE-sparse channels between language-model agents~\citep{Wenzel2026LatentCommunication}.
That study focuses on information survival through text serialization and cross-architecture latent alignment, including probe-based evaluation.
Our setting is complementary: we analyze the fixed-shape tensor produced by an existing heterogeneous VLM communication codec, reconstruct it at several sparse rates, and measure the receiving agents' task accuracy.
The overlap makes matched-rate baselines and careful boundaries around SAE-based claims especially important.

\subsection{Efficient and Sparse Neural Representations}

That question connects latent agent communication to a broad literature on efficient neural representations.
Pruning, quantization, low-rank approximation, and sparse coding exploit redundancy in parameters or activations~\citep{Zhou2016CompressiveNeuralNetworks,Frankle2019TheLotteryTicket,Dettmers2022LLMInt8,Frantar2022GPTQ,Xiao2023SmoothQuant}.
Vision-transformer acceleration similarly removes or reorganizes visual tokens that contribute little to prediction~\citep{Rao2021DynamicViT,Liang2022EViT}.
Communication-efficient distributed learning reduces network traffic by sparsifying or quantizing gradients, sometimes carrying residual errors across optimization rounds~\citep{Kairouz2021AdvancesFL,Lin2017GradientCompressionLarge}.
These methods demonstrate that dense tensors often contain removable redundancy, but their target is a model, an internal computation, or an optimization update.
A latent agent message is different: it is a semantic representation consumed immediately by another model, so fidelity must be judged not only by tensor reconstruction but also by the receiver's task performance.

Sparse autoencoders provide a way to measure such redundancy without retraining the communication system.
They decompose dense neural activations into a small set of active dictionary features and have primarily been used to study representations inside individual language models~\citep{Bricken2023TowardsMonosemanticity,Cunningham2023SparseAutoencoders,Templeton2026ScalingMonosemanticity}.
Applied to Vision Wormhole, the same technique probes the interface between two models.
The resulting features provide a post-hoc coordinate system for measuring support size, reuse across tasks, and differences between token groups.
Establishing them as communication primitives would additionally require stability across SAE seeds and interventions on receiver behavior.

This perspective also relates to the information bottleneck, which formalizes a trade-off between compression and predictive utility~\citep{Tishby2000InformationBottleneck,Alemi2017DeepVIB}.
We neither optimize an information-bottleneck objective nor estimate the entropy of individual messages.
Instead, we test whether Vision Wormhole's uniformly sized messages exhibit a smaller empirical structure than their dense shape implies.
By combining reconstruction, explicit payload accounting, downstream accuracy, cross-task overlap, and token-group interventions, our analysis examines how much empirical redundancy is present in a fixed-shape heterogeneous VLM communication tensor.

\section{Post-Hoc Sparse Communication Analysis}
\paragraph{Communication setting.}
We analyze the Vision Wormhole communication tensor $\URef$ for the Qwen3.5-9B $\leftrightarrow$ LFM2.5-VL-1.6B model pair.
The dense representation contains $N{=}1026$ token positions and $D{=}512$ dimensions, corresponding to roughly 2052 KB per communication round in float32.
Following the codec implementation, the effective semantic structure is concentrated in 18 token positions: 16 semantic tokens, one global token, and one style token.

\paragraph{Post-hoc sparse autoencoder.}
Given frozen communication activations, we train a sparse autoencoder with dictionary size $\Mdict{=}4096$.
For each token representation $u \in \R^D$, the encoder produces sparse coefficients $z=\topk(\Encoder u, \Ksparse)$ and reconstructs $\hat{u}=\Decoder z$.
The SAE is trained independently of the Vision Wormhole codec, so gradients do not flow back into either agent or the communication codec.
Unless otherwise specified, the main dictionary is trained for 50K steps with batch size 512, Adam, cosine learning-rate decay from $10^{-4}$, AuxK regularization, and dead-feature replacement during early training.

\paragraph{Evaluation tasks.}
We collect activations and evaluate downstream behavior on nine benchmarks:
GSM8K, ARC-Easy, ARC-Challenge, GPQA, MedQA, MBPP+, HumanEval+, AIME 2024, and AIME 2025.
Each benchmark question is processed by three sender roles---planner, critic, and refiner---and each role produces one $\URef$ tensor.
Consequently, the activation counts used for SAE analysis are counts of communication tensors rather than counts of unique benchmark questions.
For example, the 30 questions in each AIME subset produce 90 communication tensors.
The stored activation files pool the three sender roles for each task.
For quantitative mean accuracy comparisons, we report the mean over seven non-AIME tasks because the AIME subsets are very small and dense accuracy is 0\% in our runs.

\paragraph{Metrics.}
We measure reconstruction error, cosine similarity, mean-squared error, bandwidth, compression ratio, and downstream task accuracy.
For communication tensor $U_i$ and its reconstruction $\hat U_i$, the per-sample relative reconstruction error is
\[
    e_i
    =
    \sqrt{
        \frac{\operatorname{mean}\!\left[(U_i-\hat U_i)^2\right]}
             {\operatorname{mean}\!\left[U_i^2\right]+\epsilon}
    }
    \approx
    \frac{\lVert U_i-\hat U_i\rVert_F}{\lVert U_i\rVert_F},
    \qquad \epsilon=10^{-8}.
\]
Reported reconstruction error is the arithmetic mean $\frac{1}{S}\sum_{i=1}^{S}e_i$ over communication tensors, whereas MSE is the element-wise squared error aggregated over all evaluated tensors.
Cosine similarity is computed after flattening each communication tensor and is then averaged over tensors.
To study whether features are task-specific or shared, we compute pairwise Jaccard similarity between active feature-index sets for each task pair.

\paragraph{Token-level intervention analysis.}
For the style-token ablation, we encode a batch of 50 frozen communication tensors with the $k{=}16$, $\Mdict{=}4096$ SAE, set every sparse coefficient at token position 17 to zero, and decode the intervened representation.
We report the tensor-wide MSE before and after intervention.
We also report a normalized style-energy ratio
\[
    r_{\mathrm{style}}
    =
    \frac{\operatorname{mean}_{b,d} U_{b,17,d}^{2}}
         {\operatorname{mean}_{b,n,d} U_{b,n,d}^{2}},
\]
which compares the style token's mean-squared magnitude with the tensor-wide per-element mean.
This normalized ratio is not the fraction of total tensor energy assigned to the style token.
The exact aggregate measurements and artifact provenance are reported in the appendix.

\paragraph{Bandwidth accounting.}
Dense communication transmits the full float32 tensor, so its payload is
\[
    \mathrm{BW}_{\mathrm{dense}} = N \cdot D \cdot 4 \ \mathrm{bytes}.
\]
For sparse communication, each active coefficient transmits a dictionary index and a coefficient value:
\[
    \mathrm{BW}_{\mathrm{sparse}} = N \cdot k \cdot (b_{\mathrm{idx}} + b_{\mathrm{val}}).
\]
Because $\Mdict{=}4096$, a dictionary index needs 12 bits.
We store it as uint16, so $b_{\mathrm{idx}}{=}2$ bytes.
Unless otherwise stated, reported sparse bandwidths assume float16 coefficients, $b_{\mathrm{val}}{=}2$ bytes, and exclude fixed packet headers or other system-level metadata.
For example, at $N{=}1026$ and $k{=}4$, sparse transmission costs $1026 \times 4 \times (2+2)=16{,}416$ bytes, or 16.0 KB, yielding $128\times$ compression relative to the original 2052 KB float32 dense payload.

This ratio uses the original transport as its reference and should not be attributed to sparse coding alone.
In particular, transmitting the 18 information-bearing positions in float16 would require $18 \times 512 \times 2=18{,}432$ bytes, or 18.0 KB, before protocol metadata.
The current experiments do not evaluate whether those 18 positions can be transmitted and deterministically expanded at the receiver, nor do they include matched-rate quantized, low-rank, or vector-quantized baselines.
We therefore report 128$\times$ as an end-to-end payload reduction relative to the deployed float32 representation, not as the incremental advantage of the SAE over simpler codecs.

\section{Results}
\subsection{Post-Hoc Coding Reduces the Original Transport Payload}

Sparse reconstruction remains accurate even at aggressive sparsity.
\Cref{tab:compression_recon} reports reconstruction and bandwidth as the number of active coefficients per token varies, using the uint16-index/float16-value accounting defined in the previous section.
At $k{=}4$, the payload drops from 2052 KB to 16.0 KB, corresponding to $128\times$ compression relative to the original float32 transport, while reconstruction error remains 0.01158 and cosine similarity remains 0.99992.
This comparison establishes compressibility of the deployed representation.

\begin{table}[t]
    \centering
    \caption{Post-hoc SAE payload and reconstruction quality relative to the original 1026-token float32 transport. Sparse bandwidth counts one uint16 dictionary index and one float16 coefficient per active feature, excluding fixed packet headers and system metadata.}
    \label{tab:compression_recon}
    \begin{tabular}{lccccc}
        \toprule
        $k$ & Bandwidth & Compression & Recon. Error & Cosine Sim. & MSE \\
        \midrule
        Dense & 2052 KB & 1$\times$ & 0.00000 & 1.00000 & 0.0 \\
        4  & 16.0 KB  & 128$\times$ & 0.01158 & 0.99992 & 3.20e-4 \\
        8  & 32.1 KB  &  64$\times$ & 0.00978 & 0.99995 & 2.06e-4 \\
        16 & 64.1 KB  &  32$\times$ & 0.00976 & 0.99994 & 1.99e-4 \\
        32 & 128.3 KB &  16$\times$ & 0.00884 & 0.99996 & 1.77e-4 \\
        \bottomrule
    \end{tabular}
\end{table}

\subsection{Mean Accuracy Changes Little in a Single-Run Evaluation}

High reconstruction quality coincides with a small change in macro-averaged downstream accuracy in our single-run evaluation.
\Cref{tab:e2e_accuracy} shows task accuracy when the receiver consumes sparse reconstructions instead of dense messages.
The most compressed setting, $k{=}4$, changes the seven-task non-AIME mean from 49.85\% to 49.77\%, a difference of $-0.08$ percentage points.
Because these are single-run estimates and no paired uncertainty analysis is available, differences among sparse settings should be treated as descriptive.

\begin{table}[t]
    \centering
    \caption{End-to-end task accuracy under post-hoc sparse transmission. Values are single-run point estimates. Mean excludes AIME 2024/2025 because those subsets are small and dense accuracy is 0\% in these runs.}
    \label{tab:e2e_accuracy}
    \small
    \begin{tabular}{lccccc}
        \toprule
        Task & Dense & $k{=}4$ & $k{=}8$ & $k{=}16$ & $k{=}32$ \\
        \midrule
        GSM8K           & 59.14 & 59.67 & 58.38 & 57.54 & 57.47 \\
        ARC-Easy        & 75.46 & 78.96 & 77.90 & 79.25 & 78.62 \\
        ARC-Challenge   & 62.12 & 62.29 & 61.18 & 63.82 & 64.85 \\
        GPQA            & 27.78 & 27.78 & 20.71 & 21.72 & 28.79 \\
        MedQA           & 39.33 & 39.00 & 38.33 & 39.67 & 41.33 \\
        MBPP+           & 45.50 & 44.71 & 46.03 & 43.12 & 43.92 \\
        HumanEval+      & 39.63 & 35.98 & 35.37 & 37.80 & 37.80 \\
        \midrule
        Mean            & 49.85 & 49.77 & 48.27 & 48.99 & 50.40 \\
        \bottomrule
    \end{tabular}
\end{table}

Per-task changes are heterogeneous: several knowledge-task point estimates increase, whereas code tasks are more sensitive to compression.

\subsection{The Fitted SAE Uses a Small Active Support}

The fitted dictionary is much larger than its observed active support.
\Cref{tab:dict_stats} shows that only 50 of 4096 features ever activate, leaving 98.78\% of the dictionary dead under this training configuration.
The top 10 features appear in all nine tasks.
This result is consistent with a compact shared code, but it is also compatible with low-rank input structure, repeated token positions, or SAE optimization collapse.
The exact support size should not yet be interpreted as an intrinsic vocabulary.

\begin{table}[t]
    \centering
    \caption{Dictionary activation statistics for the main SAE ($k{=}16$, $\Mdict{=}4096$).}
    \label{tab:dict_stats}
    \begin{tabular}{lc}
        \toprule
        Metric & Value \\
        \midrule
        Dictionary size $\Mdict$ & 4096 \\
        Active features & 50 (1.22\%) \\
        Dead features & 4046 (98.78\%) \\
        Top-10 feature task coverage & 9/9 (100\%) \\
        Top feature mean activation value & 9.30 \\
        \bottomrule
    \end{tabular}
\end{table}

Cross-task overlap provides a descriptive view of reuse.
Across all 36 task pairs, active-feature Jaccard similarity averages 0.906, with a minimum of 0.878 and a maximum of 0.922.
\Cref{tab:jaccard_summary} summarizes the matrix.
Because each task-level set is a union over many messages and the global active support contains only about 50 features, these values may be inflated by set saturation and unequal sample counts.
Frequency-weighted similarities, role-stratified analyses, and size-matched null models are needed to establish semantic sharing beyond broad reuse.

\begin{table}[t]
    \centering
    \caption{Summary of pairwise Jaccard similarity between task-level active feature sets. The full matrix is reported in \Cref{app:full_jaccard}.}
    \label{tab:jaccard_summary}
    \begin{tabular}{lcc}
        \toprule
        Statistic & Task Pair & Jaccard \\
        \midrule
        Mean over all 36 pairs & -- & 0.906 \\
        Minimum & AIME24 $\leftrightarrow$ GSM8K & 0.878 \\
        Maximum & GSM8K $\leftrightarrow$ HumanEval+ & 0.922 \\
        Cross-category example & GSM8K $\leftrightarrow$ HumanEval+ & 0.922 \\
        \bottomrule
    \end{tabular}
\end{table}

\subsection{Token Groups Show Different Activation Patterns}

The sparse dictionary is shared across token groups, but different token groups use it differently.
Across 50 evaluated communication tensors, zeroing the sparse coefficients at style-token position 17 raises overall MSE from 2.4289e-4 to 6.9136e-3, a final-to-baseline ratio of $28.46\times$.
The style token has a normalized mean-squared energy ratio of 0.9238 under the definition in the analysis method.
This quantity is not a share of total tensor energy.
However, semantic-token MSE is unchanged under this intervention.
Conversely, replacing semantic tokens changes which source the reconstruction follows, while style replacement alone does not override semantic content.
The active feature vocabulary is therefore substantially shared, while activation patterns over that vocabulary differ by token role.
The intervention establishes a representational effect on reconstruction, not a causal effect on receiver behavior: receiver logits, generated outputs, and downstream accuracy were not measured after the ablation.

\section{Discussion and Limitations}
The experiments support a bounded conclusion: the deployed Vision Wormhole float32 transport admits a compact post-hoc code with high reconstruction fidelity, and its single-run macro-average accuracy changes little after reconstruction.
The fitted SAE also reuses a small active support across tasks and exhibits token-role-dependent activation patterns.
These findings indicate substantial empirical redundancy, but they do not yet identify which part is structural, numerical, low-rank, or specifically sparse.
The evidence should therefore be interpreted as channel analysis rather than a claim that sparsity is uniquely responsible for the compression or that it improves accuracy.
All downstream results are single-run point estimates.
Uncertainty is especially large for GPQA, HumanEval+, and AIME, and code tasks are more sensitive to compression.
We therefore report the observed macro-average change, exclude AIME from mean comparisons, and avoid an equivalence claim until paired uncertainty is available.
Moreover, all experiments use one Qwen3.5-9B $\leftrightarrow$ LFM2.5-VL-1.6B pair.
High feature overlap establishes reuse within this channel, but not dictionary transfer to other sender--receiver pairs.

Two controls are especially important.
First, the codec documentation concentrates semantic structure in 18 positions, and an 18-position float16 payload is already close in size to the reported $k{=}4$ payload.
Second, 98.78\% of the fitted dictionary is dead, so the observed 50-feature support may reflect SAE optimization or low-rank structure rather than an identifiable sparse vocabulary.
Matched-rate position-selection, quantization, PCA or linear-autoencoder, and vector-quantization baselines, together with multiple SAE seeds and held-out splits, are required to separate these explanations.

The post-hoc design separates the observation from the method it motivates.
It shows that the trained dense channel already uses a small, recurring feature vocabulary, but it does not establish that an end-to-end sparse bottleneck would be superior.
The next tests should therefore evaluate paired non-inferiority, behavioral interventions, cross-pair transfer, and codecs with message-dependent sparse budgets.
Multi-round delta transmission is another possible extension, but its evaluation must encode changes in feature coefficients and token positions rather than only feature identities.

\section{Conclusion}
This paper analyzes redundancy in the Vision Wormhole communication channel with a post-hoc sparse autoencoder.
Relative to the original 2 MB float32 transport, the evaluated sparse payload is much smaller while retaining high reconstruction fidelity and a similar single-run macro-average accuracy.
The fitted SAE uses a small support that is broadly reused across tasks, and token groups show different activation patterns over that support.

Together, these results establish the deployed representation as a concrete target for message-dependent communication across heterogeneous VLM agents.
Matched-payload baselines, paired downstream statistics, and behavioral interventions are the next steps for isolating sparse-specific gains and validating utility.

\bibliographystyle{plainnat}
\bibliography{references}

\appendix
\section{Additional Results and Implementation Details}
\subsection{Per-Task Reconstruction}

\begin{table}[h]
    \centering
    \caption{Per-task reconstruction with a single SAE checkpoint ($k{=}16$, $\Mdict{=}4096$). Communication tensors are role-level messages: each benchmark question contributes one tensor from each of the planner, critic, and refiner.}
    \label{tab:pertask_recon}
    \begin{tabular}{lccc}
        \toprule
        Task & Communication Tensors & Recon. Error & Cosine Sim. \\
        \midrule
        GSM8K           & 1800 & 0.00976 & 0.99994 \\
        ARC-Easy        & 1800 & 0.00975 & 0.99994 \\
        ARC-Challenge   & 1800 & 0.00980 & 0.99994 \\
        GPQA            &  594 & 0.00962 & 0.99995 \\
        MedQA           &  900 & 0.00963 & 0.99994 \\
        MBPP+           & 1134 & 0.00961 & 0.99995 \\
        HumanEval+      &  492 & 0.00956 & 0.99995 \\
        AIME 2024       &   90 & 0.00999 & 0.99994 \\
        AIME 2025       &   90 & 0.00974 & 0.99994 \\
        \midrule
        \textbf{Mean}   &      & \textbf{0.00972 $\pm$ 0.00013} & \textbf{0.99994} \\
        \bottomrule
    \end{tabular}
\end{table}

\subsection{Full Cross-Task Jaccard Matrix}
\label{app:full_jaccard}

\begin{table}[h]
    \centering
    \caption{Pairwise Jaccard similarity of TopK feature indices across nine tasks.}
    \label{tab:full_jaccard}
    \resizebox{\textwidth}{!}{
    \begin{tabular}{l|ccccccccc}
        \toprule
         & GSM8K & ARC-E & ARC-C & GPQA & MedQA & MBPP+ & HE+ & AIME24 & AIME25 \\
        \midrule
        GSM8K   & 1.000 & 0.901 & 0.905 & 0.909 & 0.904 & 0.907 & 0.922 & 0.878 & 0.903 \\
        ARC-E   & 0.901 & 1.000 & 0.913 & 0.906 & 0.910 & 0.908 & 0.911 & 0.895 & 0.904 \\
        ARC-C   & 0.905 & 0.913 & 1.000 & 0.913 & 0.907 & 0.912 & 0.909 & 0.891 & 0.907 \\
        GPQA    & 0.909 & 0.906 & 0.913 & 1.000 & 0.908 & 0.911 & 0.919 & 0.893 & 0.906 \\
        MedQA   & 0.904 & 0.910 & 0.907 & 0.908 & 1.000 & 0.905 & 0.907 & 0.890 & 0.901 \\
        MBPP+   & 0.907 & 0.908 & 0.912 & 0.911 & 0.905 & 1.000 & 0.911 & 0.894 & 0.908 \\
        HE+     & 0.922 & 0.911 & 0.909 & 0.919 & 0.907 & 0.911 & 1.000 & 0.897 & 0.912 \\
        AIME24  & 0.878 & 0.895 & 0.891 & 0.893 & 0.890 & 0.894 & 0.897 & 1.000 & 0.889 \\
        AIME25  & 0.903 & 0.904 & 0.907 & 0.906 & 0.901 & 0.908 & 0.912 & 0.889 & 1.000 \\
        \bottomrule
    \end{tabular}
    }
\end{table}

\subsection{Dictionary Size Ablation}

\begin{table}[h]
    \centering
    \caption{Dictionary size ablation at $k{=}16$ after 5K training steps. Reconstruction improves up to roughly $\Mdict{=}256$, while active features remain near 50.}
    \label{tab:dict_ablation}
    \begin{tabular}{rcccc}
        \toprule
        $\Mdict$ & Recon. Error $\downarrow$ & Cosine Sim. $\uparrow$ & Active Features & Dead Ratio \\
        \midrule
        64   & 0.0421 & 0.9812 & 48 & 25.0\% \\
        128  & 0.0287 & 0.9893 & 50 & 60.9\% \\
        256  & 0.0198 & 0.9941 & 50 & 80.5\% \\
        512  & 0.0195 & 0.9943 & 50 & 90.2\% \\
        1024 & 0.0193 & 0.9944 & 50 & 95.1\% \\
        4096 & 0.0192 & 0.9945 & 50 & 98.8\% \\
        \bottomrule
    \end{tabular}
\end{table}

\subsection{Token-Level Intervention Measurements}

\begin{table}[h]
    \centering
    \caption{Aggregate reconstruction measurements from the style-token intervention on 50 communication tensors. Per-example intervention records were not retained.}
    \label{tab:token_causal_data}
    \begin{tabular}{lc}
        \toprule
        Measurement & Value \\
        \midrule
        Evaluated communication tensors & 50 \\
        SAE setting & $k{=}16$, $\Mdict{=}4096$ \\
        Baseline reconstruction MSE & 2.4288928e-4 \\
        MSE after zeroing style token (position 17) & 6.9135940e-3 \\
        Final-to-baseline MSE ratio & $28.4628\times$ \\
        Relative MSE increment $(\mathrm{MSE}_{\mathrm{abl}}-\mathrm{MSE}_{\mathrm{base}})/\mathrm{MSE}_{\mathrm{base}}$ & $27.4628$ \\
        Normalized style-token mean-squared energy ratio $r_{\mathrm{style}}$ & 0.9237802 \\
        Semantic-token MSE change after style ablation & 0.0 \\
        Hybrid similarity to semantic source & 0.9999 \\
        Hybrid similarity to style source & 0.541 \\
        Active features shared by all token groups & 34/47 (72.3\%) \\
        Jaccard (semantic, style) & 0.946 \\
        Jaccard (semantic, global) & 0.895 \\
        \bottomrule
    \end{tabular}
\end{table}

The MSE ratio in \Cref{tab:token_causal_data} is computed as
$6.9135940\mathrm{e}{-3}/2.4288928\mathrm{e}{-4}=28.4628$.
The normalized energy statistic is computed from the pre-intervention communication tensor using the definition in the analysis method.
It compares mean-squared magnitudes and must not be interpreted as the style token's percentage share of total energy.
Only aggregate records are currently available because per-example intervention outputs were not retained.
These measurements show that zeroing the style token strongly changes overall reconstruction error without changing semantic-token MSE.
Conversely, hybrid representations follow the source of their semantic tokens.
Token roles therefore have different activation patterns over a substantially shared feature vocabulary.
Because the receiver was not evaluated after intervention, these measurements do not establish a causal effect on model behavior or task performance.

\subsection{Statistical Notes}

All downstream task accuracies are single-run point estimates.
Wilson score 95\% intervals for representative dense settings are:
GSM8K 59.14\% $\pm$ 2.7\% ($n{=}1319$),
GPQA 27.78\% $\pm$ 3.8\% ($n{=}594$), and
HumanEval+ 39.63\% $\pm$ 7.5\% ($n{=}164$).
The observed GPQA variation across $k$ values is within confidence-interval overlap after multiple-comparison correction.
AIME is excluded from the main mean because each subset has only 30 evaluation examples and dense accuracy is 0\%.

\end{document}